\documentclass{article}

\PassOptionsToPackage{numbers, compress}{natbib}

\usepackage[preprint]{neurips_2020}

\usepackage[utf8]{inputenc} 
\usepackage[T1]{fontenc}    
\usepackage{hyperref}       
\usepackage{url}            
\usepackage{booktabs}       
\usepackage{amsfonts}       
\usepackage{nicefrac}       
\usepackage{microtype}      
\usepackage{amsmath}
\usepackage{graphicx}
\usepackage{comment}
\usepackage{multirow}
\usepackage{xcolor}
\usepackage[title]{appendix}

\title{Background Splitting: \\Finding Rare Classes in a Sea of Background}

\author{%
  Ravi Teja Mullapudi*\\
  Carnegie Mellon University\\
  \texttt{rmullapu@cs.cmu.edu}\\
  \And
  Fait Poms*\\
  Stanford University\\
  \texttt{fpoms@cs.stanford.edu}\\
  \And
  William R. Mark\\
  Google Research\\
  \texttt{billmark@google.com}\\
  \And
  Deva Ramanan\\
  Carnegie Mellon University\\
  \texttt{deva@cs.cmu.edu}\\
  \And
  Kayvon Fatahalian\\
  Stanford University\\
  \texttt{kayvonf@cs.stanford.edu}\\
}

\begin{document}

\maketitle
\let\thefootnote\relax\footnote{*Both authors contributed equally to this paper.}
\begin{abstract}
 
We focus on the real-world problem of training accurate deep models for image classification of a small number of rare categories. In these scenarios, almost all images belong to the background category in the dataset (>95\% of the dataset is background). We demonstrate that both standard fine-tuning approaches and state-of-the-art approaches for training on imbalanced datasets do not produce accurate deep models in the presence of this extreme imbalance.
Our key observation is that the extreme imbalance due to the background category can be drastically reduced by leveraging visual knowledge from an existing pre-trained model. Specifically, the background category is "split" into smaller and more coherent pseudo-categories during training using a pre-trained model. We incorporate \emph{background splitting} into an image classification model by adding an auxiliary loss that learns to mimic the predictions of the existing, pre-trained image classification model. Note that this process is automatic and requires no additional manual labels. The auxiliary loss regularizes the feature representation of the shared network trunk by requiring it to discriminate between previously homogeneous background instances and reduces overfitting to the small number of rare category positives. 
In addition, we also show that background splitting can be combined with other methods which modify the main classification loss to deal with background imbalance to further improve performance, indicating that background splitting is a complementary method to existing imbalanced learning techniques.
We evaluate our method on a modified version of the iNaturalist dataset where only a small subset of rare category labels are available during training (all other images are labeled as background). By jointly learning to recognize ImageNet categories and selected iNaturalist categories, our approach yields performance that is 42.3 mAP points higher than a standard fine-tuning baseline when 99.98\% of the data is background, and 8.3 mAP points higher than state-of-the-art baselines when 98.30\% of the data is background.



\end{abstract}

\section{Introduction} 

\begin{figure}[t]
\includegraphics[width=\textwidth]{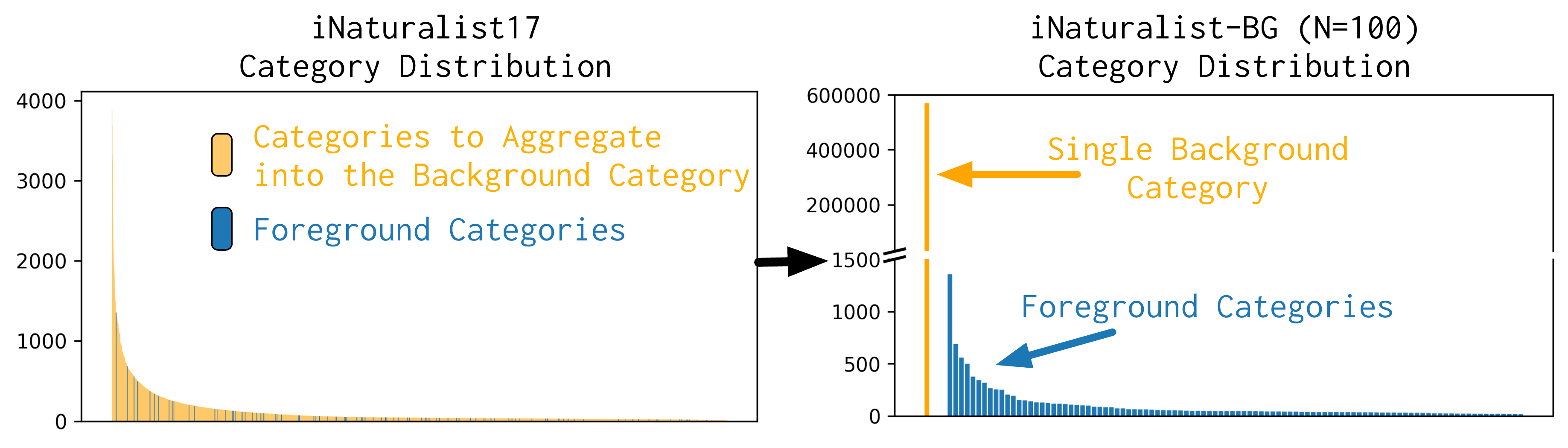}
\caption{Left: distribution of examples for iNaturalist, sorted by category frequency (y-axis).
Although there is no background category in iNaturalist,
we color examples based on whether they are placed in the background category in our modified dataset, iNaturalist-BG.
Right: distribution of examples in iNaturalist-BG ($N\!=\!100$).
Notice that the background category accounts for most (98.3\%) of the examples in the dataset.}
\label{fig:overview}
\end{figure}

Image classification tends to be evaluated on datasets that have been manually curated to be clean and balanced~\cite{russakovsky15imagenet, lin2014microsoft}.
Recent focus has shifted toward in-the-wild data that features long-tail distributions~\cite{gvanhorn2018inat, salakhutdinov2011learning, wang2016learning, wang2017learning, zhu2014capturing}, 
but these datasets are still ``artificial'' in that every curated image is a positive example of some category\,\cite{liu2020oltr,liu2020oltr,cao2019learning,cao2019learning,gvanhorn2018inat}. 
In contrast, in many real-world settings, categories of interest are sufficiently rare that it is far more common for images to \emph{not exhibit any categories of interest}.
For example, consider a newly collected dataset (e.g., obtained by an autonomous fleet) where a new category -- an e-scooter -- is annotated. This is essentially a binary image classification problem that is highly imbalanced and contains only sparse positives, as the vast majority of collected images will {\em not} contain an e-scooter.
If additional categories were desired, one can model the task as a multi-way image classification problem where a significant fraction of images are labeled as the ``background'' category. 



This work focuses on the problem of training accurate deep models for image classification of a small number of rare categories. In these scenarios, not only is there a small number of positives per category, but the overall number of positive examples for all categories is dominated by the number of background images in the dataset (>95\% of the dataset is background).
We find that state-of-the-art methods for deep imbalanced classification based on data sampling~\cite{kang2020decoupling} or loss reweighting~\cite{cao2019learning} struggle in this setting.

{\bf Contributions:} 
We propose a new and surprisingly simple method to address severe background imbalance:
reduce the imbalance by conceptually \emph{splitting the background category} into many smaller categories during training.
Rather than modify the main (rare category) classification loss of the model to identify more categories, we add an auxiliary loss that learns to mimic the predictions of an existing,
pre-trained image classification model that recognizes more commonly occurring categories.
Therefore, our approach transfers knowledge of an auxiliary classification task into the target model via distillation---it only requires access to the pre-trained model, \emph{not access to any additional labels} beyond those used for the primary rare category classification task.
By jointly learning the main classification loss and the auxililary classification loss, 
we reduce dominance of the background category and also avoid overfitting to the small number of rare category positives.
We also show that our auxiliary classification loss can be combined with other methods for dealing with a large background category. Specifically, we show that treating the background category as a default that is chosen when no other category is strongly indicated, following Matan et al.~\cite{matan1990handwritten}, is complementary to background splitting.


{\bf Benchmarking:} 
We benchmark in the large background category setting by modifying the label distribution of the highly-imbalanced, large-vocabulary iNaturalist dataset~\cite{gvanhorn2018inat}. We selected iNaturalist 2017 because it is a very large dataset (675,170 images across both the training and validation set) with a large category vocabulary (5089 categories) and is based on a real-world use case of identifying the world’s flora and fauna, collected by thousands of real individuals, instead of artificially choosing a list of categories and collecting images by querying internet search engines~\cite{liu2020oltr, cao2019learning}.


We modify the iNaturalist label distribution by aggregating multiple categories together into a single background category, reassigning images labeled for those categories to the background label. We call this modified dataset \textbf{iNaturalist-BG}. We created \textbf{iNaturalistBG} to mimic the real-world scenario in which a large unlabeled dataset has been collected and then labeled for a small number of categories. We believe this is an important and common use case that is under served by existing datasets. 
This dataset provides several different train and test set pairs to study how an increasingly large background category affects model performance.  We denote the train and test set pairs using the number of foreground (non-background) categories $N$ that they contain. For example, Figure~\ref{fig:overview}-right shows the distribution of images to categories for the $N\!=\!100$ training set. Note that in all cases we train on the full 579,184 images in the iNaturalist training set, where images not belonging to the $N$ categories of interest are labelled with the background class.

{\bf Analysis:} 
We find that state-of-the-art techniques for imbalanced classification~\cite{kang2020decoupling,cao2019learning} perform \emph{worse} than standard fine-tuning in scenarios where the background category makes up more than 98\% of the dataset (corresponding to $N\!=\!100$). In contrast, our method of background splitting outperforms prior art by 8.9 mAP points. In more extreme single foreground category setting ($N\!=\!1$, corresponding to 99.98\% background labels), our method provides a staggering mAP improvement from 10.6 to 52.9.

We plan to release our code and the code for generating the \textbf{iNautralistBG} dataset shortly.

\section{Related Work}

Given our problem formulation, there are three key design questions:  (1) How do we address category imbalance? (2) How do we address diversity in the background category? and (3) How can we leverage useful information from a related dataset such as ImageNet?  We discuss related work in each of these areas. 

\paragraph{Category imbalance.}
Most methods for learning under long-tail, imbalanced datasets~\cite{he2009learning} use one (or more) of the following techniques: re-balancing, re-weighting/category-conditioned adjustments to final loss values, and category clustering.
\textbf{Re-balancing methods} alter the training distribution to simulate traditional balanced training sets \cite{cao2019learning, van2017devil}.
Recently it has been found that re-balancing should be used only in a final stage of model training to encourage more general feature representations~\cite{kang2020decoupling, cao2019learning, wang2019classification}.  Object detectors, which suffer from severe foreground-to-background imbalance~\cite{oksuz2020imbalance}, use re-sampling techniques such as hard negative mining \cite{shrivastava2016training}.
\textbf{Re-weighting/category-conditioned methods} adjust the loss attributed to a sample based on its category~\cite{zhu2020alleviating, cao2019learning, kang2020decoupling} or how hard the sample is.  Object detectors rely heavily on these methods~\cite{lin2017focal}, where they are particularly useful because the effective number of samples contributing to a batch's loss is much larger than in image classifiers . 
\textbf{Category clustering methods} use clustering of embeddings from training samples to help improve transfer from head to tail categories \cite{liu2020oltr} and to train sub-models specialized to individual categories (avoiding imbalance) \cite{zhu2014capturing, ouyang2016factors, li2020overcoming}.


{\bf Background category.}  An important special case of imbalanced classification, and the focus of the iNaturalistBG dataset we introduce in this paper, occurs when there is a highly {\it diverse} background category that is far more common than any other foreground category. In this setting, models are given a set of images $D_c, D_b$ belonging to the foreground categories and background category, respectively, during training, and are expected to classify $D_c', D_b'$ at test time. Much existing work has focused on how to best train classifiers that can separate $D_c$ from $D_b$ by: adding an additional "N+1" category that represents the background~\cite{matan1990handwritten, loncaric2018background, liu2016ssd}, modeling the background as a Gaussian distribution~\cite{osadchy2012hybrid}, or by training the network to predict a low scores for background instances and using a threshold~\cite{dhamija2018reducing, matan1990handwritten}. This problem is related to the foreground-background class imbalance problem in object detection~\cite{oksuz2020imbalance}, though existing long-tail object detection datasets with rare categories~\cite{gupta2019lvis, OpenImages2} only provide a relatively small set of negative annotations for each rare category (typically less than 1000), and so do not provide a setting for evaluating significant background imbalance. In this paper, we study the background imbalance problem in isolation in the context of image classification.

\paragraph{Open World, Open Set, and Out-of-distribution recognition.} Unlike the background category setting (in which models are given access to train instances $D_b$ which are from the same distribution as the test instances $D_b'$), Open World, Open Set, or Out-of-distribution recognition datasets test on data from categories (Open World/Open Set) or entire datasets (Out-of-distribution) not seen during training~\cite{liu2020oltr, bendale2015towards, bendale2016towards, dhamija2018reducing}. In this setting, the model is tasked with identifying "unknown unknowns" in the test distribution which were not present in the training distribution. Typically, these "unknown unknowns" are not a significant majority of the test distribution. In contrast, in our setting we are interested in "known unknowns" which are present in the training distribution ($D_b$) and where these "known unknowns" are a majority of the data (>95\%).

{\bf Knowledge transfer and sharing.}  There are a variety of ways in which knowledge from one task or dataset can be used to help train a network for a related task or dataset.  {\bf Transfer learning} explicitly focuses on this problem~\cite{pan2009survey,razavian2014cnn, shao2015transfer, kornblith2018transfer}.  The most common form of it is to pre-train a model on the first task, then fine-tune it on the target task with uniform sampling and cross entropy loss. However, in the presence of a dominant background category this technique results in models that focus on the easy background examples and ignore the rare positives. Our approach modifies the standard transfer learning by introducing an auxiliary distillation loss to improve model performance in these challenging settings. The auxiliary loss ensures that learned feature representation can discriminate between the fine-grained foreground categories and the few similar but hard negative examples that exist in the large set of background examples.
\textbf{Multi-task learning} can also be be viewed as a form of knowledge sharing where jointly training for a set of tasks results in better models than training a model for each individual task~\cite{caruana1997multitask, lee2013pseudo}.  Our approach is similar in spirit and is also is inspired by multi-task learning approaches to incremental learning~\cite{li2017learning,silver2002task,klingner:2020,shmelkov:2017}. However, our approach uses supervision from a existing classifier to regularize training rather than trying to train a model that accurately classifies both prior and novel categories.  {\bf Knowledge distillation.}  Instead of transferring knowledge using fine-tuning, as is common in transfer learning, we transfer knowledge using a form of knowledge distillation \cite{hinton2015distilling, wang2020knowledge}, where the distillation is performed using data from the new domain rather than data from the original domain .  This approach to knowledge transfer has been shown to outperform fine-tuning for some tasks \cite{su2017adapting, liu2019semanticaware}, and is also useful when the source and destination models are different \cite{girdhar2019distinit}.




%
%

\paragraph{Choice of baselines, focusing on category imbalance.}  The state-of-the-art baselines for training on the iNaturalist dataset are \textbf{LWS}~\cite{kang2020decoupling} and \textbf{LDAM}~\cite{cao2019learning}. These two methods can be seen as representing the main high-level approaches for dealing with imbalanced data: \textbf{LWS} uses category-balanced sampling, while \textbf{LDAM} uses a loss re-weighting approach.  We compare our method to these baselines as well as to standard fine-tuning with a cross entropy loss.  
\section{Method}

\begin{figure}[t]
\includegraphics[width=\textwidth]{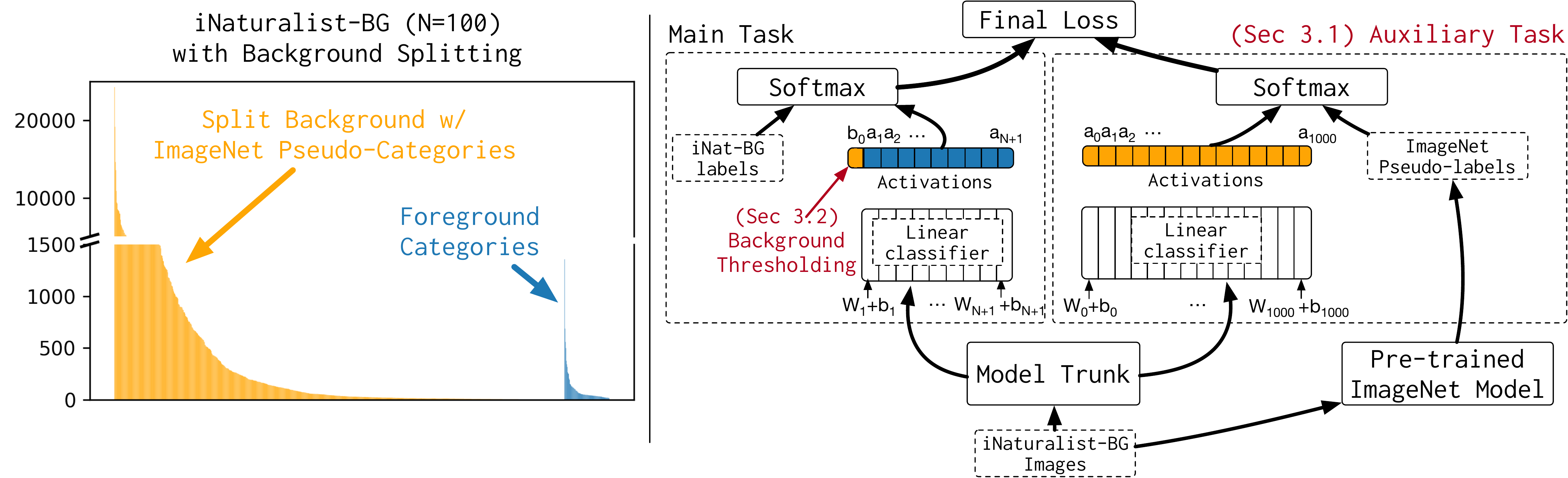}
\caption{Left: distribution of examples for the $N\!=\!100$ iNaturalist-BG dataset sorted by category where the background is broken into automatically assigned pseudo-categories using a pre-trained ImageNet model. Right: a visualization of the model graph for our method. We use a multi-task loss that is supervised with iNaturalist-BG labels and pseudo-labels generated by a pre-trained ImageNet model. The main task loss uses a fixed background logit to improve background classification.}
\label{fig:method}
\end{figure}



We address the challenges in training models with a sparse number of positives and a large background category by transforming the optimization problem into a easier one with significantly lower distribution skew. We also describe how we combine our background splitting technique with the common softmax thresholding method introduced by Matan et al.~\cite{matan1990handwritten}.

\subsection{Regularization with an auxiliary task} 

\begin{figure}[t]
\includegraphics[width=\textwidth]{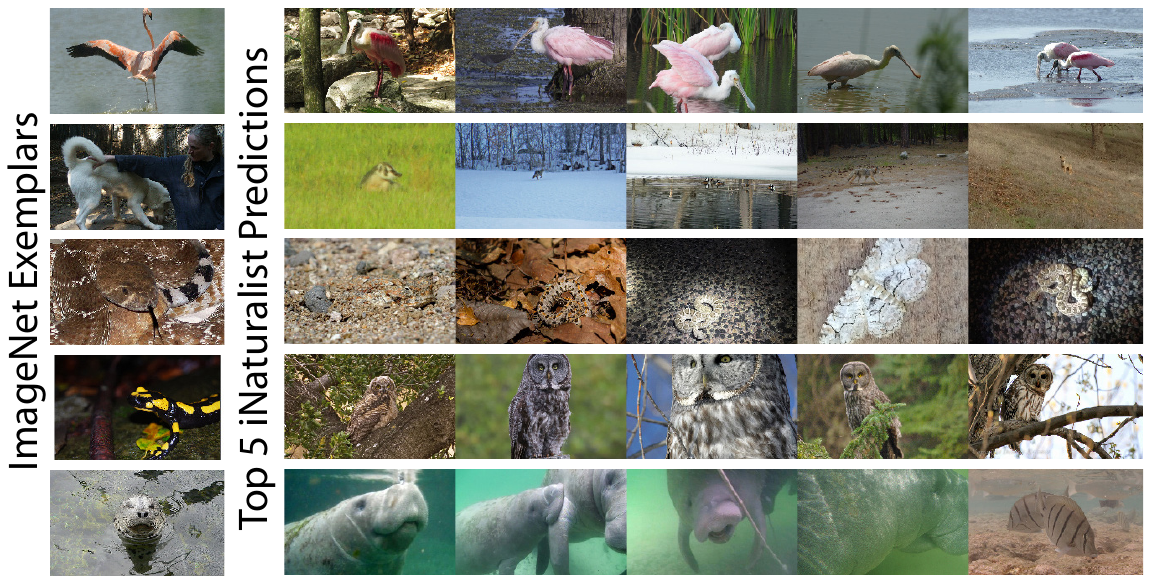}
\caption{A model pre-trained on ImageNet groups semantically similar images together in iNaturalist, resulting in pseudo-labels that effectively split the large background category in iNaturalistBG. Each row represents a category chosen at random from the ImageNet categories: the image on the left is an example image from the category taken from ImageNet, and the images on the right are the top 5 most confident predictions on the iNaturalist dataset for that category made by a model pre-trained for ImageNet. Notice the visually similar images for rows 1, 3, 4, and 5 even though the ImageNet category is different from the objects in the images.}
\label{fig:auxloss}
\end{figure}


 When training the model for the task of classifying a small set of rare categories (\emph{main task}), we want a mechanism to avoid degenerate solutions which focus on the large number of background examples or overfit to the small set of foreground positives. Our solution is to effectively ``split up'' the majority background category into sub-categories to reduce the distribution skew. However, this requires additional labels on the examples in the background category. Our key idea is that we can machine generate \emph{pseudo-labels} which can be used to partition the background into sub-categories with no additional human effort. More concretely, we generate the pseudo-labels by running a model pre-trained  on a different classification dataset on all the training data. 
 
We want the background sub-categories to significantly reduce category distribution skew and encourage the network to maintain a robust feature representation by breaking up the background into many \emph{visually or semantically coherent} categories. To reduce skew, we need a model that will separate our dataset into many sub-categories. To encourage a robust feature representation, we need a model that produces sub-categories which are visually or semantically coherent (i.e., contain images which are from similar categories) so that differentiating between categories requires robust features.  Therefore, we chose a standard ResNet50 model~\cite{he2016deep} trained on the 1000-way large-scale image classification dataset ImageNet~\cite{russakovsky15imagenet} and evaluated it on every image in the iNaturalist dataset to produce the pseudo-labels. Figure~\ref{fig:method}-left shows the background category for $N\!=\!100$ split into the 1000 ImageNet categories generated by the pre-trained model. Notice how the most frequent background sub-category now contains at most 4.5\% of the dataset, instead of the prior 98.3\%, achieving our goal of a model that separates the dataset. Figure~\ref{fig:auxloss} investigates the coherence of the categories produced by the model. Notice how rows 1, 3, 4, and 5 contain visually similar (coherent) images even though they do not map exactly to the ImageNet category as seen at left (row 1's ImageNet category is "flamingo," but the birds at right are a different species of pink bird; row 4's ImageNet category is "european fire salamander," but the images to the right are of owls); it's not necessary for there to be a direct mapping between the ImageNet categories and the iNaturalist categories for coherent groupings to occur. Appendix~A provides further analysis of how the choice of method to generate the pseudo labels for the auxiliary task affects model performance, particularly when coherence within categories is \emph{not} achieved.

More concretely, in addition to supervising the model with the main task's cross-entropy loss for $(N\!+\!1)$-way classification during training (the additional $+\!1$ category representing the background), we also add an \emph{auxiliary task} by attaching a second classification head to the network supervised by the pseudo-labels that were generated with the pre-trained ImageNet model  to serve as regularization (see \emph{(Sec 3.1) Auxiliary Task} in Figure~\ref{fig:method}-right). 
Formally speaking, let $y \in \{0, \ldots N\}$ be the main $(N\!+\!1)$-way classification task where $0$ is treated as the background class. Given a training set with a large fraction of background examples, assume that we have access to a pre-trained network that provides (ImageNet) pseudo-labels on all training pairs $\{(x_i,y_i)\}$:
\begin{align}
    \{(x_i,y_i,t_i)\} \qquad y_i \in \{0, \ldots N\}, \quad t_i \in \{1, \ldots K\}.
\end{align}
where $t_i$ is the ImageNet pseudo-label.
We  then learn a classifier with a multi-task loss:
\begin{align}
    \min_{\theta,{\bf w},{\bf v}} \sum_i \text{loss}(y_i,F_{\theta,{\bf w}}(x_i)) + \lambda_G\text{loss}(t_i,G_{\theta,{\bf v}}(x_i)) 
\end{align}
\noindent where $F$ and $G$ refer to the main and auxiliary task classification heads, respectively, of a base network trunk with shared features ${\theta}$ and task-specific features ${\bf w}$ and ${\bf v}$, $\lambda_G$ is the weight for the auxiliary loss, and the loss function is the standard softmax cross-entropy loss. Figure~\ref{fig:method}-right visualizes the full training graph for our multitask network, which defines task-specific linear weights $({\bf w},{\bf v})$ on the shared feature trunk $\theta$. 


\subsection{Background thresholding} 
The straight-forward approach to performing classification with a background category 
modifies the N-way classification problem into an $(N\!+\!1)$-way classification problem.
However, linear (softmax) classification encourages examples that fall into the same category to have similar features that can be linearly separated from other classes. This may be problematic for the background class, which we expect to be very large and diverse in our setting (Figure~\ref{fig:overview}-left). 
We alleviate this issue by {\em choosing to not} learn a classifier for the background category. Instead, we adopt the method proposed by Matan et al.~\cite{matan1990handwritten}, which assigns the background category a fixed activation, as represented by $b_0$ in Figure~\ref{fig:method}-right.
We write the softmax classifier $F_{\theta,{\bf w}}(x)$ as follows:
\begin{align*}
    F_{\theta,{\bf w}}(x)[n] := p(y=n|x) 
    \propto e^{w_n \cdot f_\theta(x) + b_n} , \quad n \in \{0,1,\ldots N\} \qquad \text{[Softmax Classification]}
\end{align*}
\noindent where $f_\theta(x)$ is the nonlinear (deep) embedding shared across our heads and ${\bf w} = w_0,b_0,\ldots,w_{N+1},b_{N+1}\}$. 
Because the weight associated with the background class ($n = 0$) can be difficult to estimate given the large expected variability in appearance, we {\em clamp} it to be the 0 vector:
\begin{align*}
    w_0 = {\bf 0}, b_0 = \text{constant} \qquad \text{[Background Thresholding]}
\end{align*}
and do not update either during learning. This modification is equivalent to setting the background class logit to a fixed constant during learning:
\begin{align}
    p(y=n|x)  = \frac{e^{w_n \cdot f_\theta(x) + b_n}}{e^{b_0} + \sum_{i=1}^{N+1}  e^{w_i \cdot f_\theta(x) + b_i}}, \qquad n \in \{1,\ldots N\} \label{eq:bgmax}
\end{align}
where the sum over $N$ class probabilities can now be less than 1 (where the remaining probability mass is assigned to the background class). The above model no longer learns a hyperplane $w_0$ to separate background examples from the foreground category examples, but instead classifies an example as background only if the model produces logit values lower than $b_0$ for all foreground categories. 
\section{Evaluation}

We evaluate our method against baselines using \textbf{iNaturalist-BG}, a modified variant of the long-tailed and fine-grained iNaturalist dataset. In majority background regimes, we demonstrate that our approach yields significantly more accurate classification models compared to standard fine-tuning.
We also compare our approach with prior work on training classifiers in imbalanced settings,
and show that our approach dramatically outperforms these prior methods on our task.
Finally, we perform a factor analysis to understand how well our auxiliary loss approach works when combined with other methods for dealing with background imbalance, like the BG thresholding method.

\subsection{Experimental setup}
We simulate settings that have extreme imbalance and a dominant background category by creating a modified variant of the iNaturalist 2017 dataset, which we name iNaturalist-BG.  The original iNaturalist 2017 dataset has 579,184 training images and 95,986 validation images. The dataset is long-tailed and each individual category is rare.  In our modified dataset, a small number of categories are considered to be labelled for the entire dataset, while the remainder of the categories from the original dataset are combined into a single background category.  To study how the size of the background category affects model performance, we provide several different training and tests set pairs.  We denote the train and test pairs using the number of foreground (non-background) categories it contains: $N\!=\!5089$, equivalent to the standard iNaturalist 2017 category distribution, $N\!=\!1100$, $N\!=\!100$, $N\!=\!10$ and $N\!=\!1$.  These training and test set pairs are constructed so that performance metrics (such as mAP and F1) computed on the test sets can be compared across choices of $N$, even though the category distribution across the training and test set pairs differs. To do so, we selected a fixed set of 100 test categories which we use to evaluate performance for each of the train and test pairs.  (Since $N\!=\!10$ and $N\!=\!1$ can not contain all 100 test categories at once, we provide 10 test set pairs for $N\!=\!10$ and 100 for $N\!=\!1$ which together contain all the 100 test categories, and then average performance across these subsets). Note that in all cases we train on the full 579,184 images in the iNaturalist training set, but the percentage of images in the background changes drastically.

\begin{itemize}
\item 
$N\!=\!5098$: Train a single model with all the 5098 categories in the dataset labeled. This corresponds to the standard setting on the iNaturalist dataset, with 0\% of the data in the background class.
\item 
$N\!=\!1100$: Single model with 1100 categories with the 100 test categories included in $N$. 77.95\% of data is in the background category.
\item
$N\!=\!100$: Single model for all the 100 random test categories. 98.3\% of data is in the background category.
\item
$N\!=\!10$: 10 models each trained with a subset of 10 categories from the 100 randomly chosen categories. 99.83\% of data is in the background category on average.
\item
$N\!=\!1$: 10 binary models for a subset of 10 categories from the randomly chosen categories. We limit evaluate our evaluation to 10 categories in the case of binary models since it is too expensive to train separate deep binary models for all the 100 categories evaluated in other settings. As such, the performance for this setting is not directly comparable to the other settings.  99.98\% of data is in the background category on average.
\end{itemize}

\paragraph{Evaluation metrics.} Traditional metrics for image classification include top-1 or top-5 error, but in our setting, these can be gamed by favoring the background. Instead, we propose two evaluation protocols for a test dataset with $N$ class labels and a large background. The first protocol is motivated by top-1 error: algorithms must report a single $(N+1)$-way label for each test sample. It computes the ${\bf F1}$ accuracy (harmonic mean of precision and recall) for each of the $N$ classes and reports their mean. The second protocol, motivated by those used for large-background tasks such as object detection~\cite{everingham2015pascal}, recasts the problem as $N$ retrieval tasks corresponding to each foreground category. For each category, algorithms must return a {\em confidence} for each test sample. These are ranked to produce $N$ precision-recall curves, which in turn are summarized by their average area underneath ({\bf AP}). 

\paragraph{Model and training details.} We use the ResNet-50 architecture in all our experiments and we initialize the model from a ImageNet pre-trained ResNet-50 model. We set the background threshold value to $b_0\!=\!0.1$ and we set the weight on the auxiliary loss to $\lambda_G\!=\!0.1$.  We find that large batch sizes are crucial for training models with standard fine-tuning (\textbf{FT}) when the background category is dominant. In all our experiments we use a batch size of 1024 for \textbf{BG Splitting} and 512 for the \textbf{FT} baseline. (We found batch size 512 to perform marginally better than 1024 for \textbf{BG Splitting}.)
We provide additional results on the effect of batch size on different methods in Appendix~\ref{sec:batch_size_study}.

\paragraph{Baselines.} We compare our approach to standard fine-tuning with a cross entropy loss. (We denote this baseline as \textbf{FT} in the rest of the paper.)  We tuned the learning rate and batch size for the \textbf{FT} baseline via hyper parameter search. We also attempted to improve the \textbf{FT} baseline using common techniques such as downweighting the loss due to background samples and balancing mini-batches. We found that those common techniques did not improve performance of the baseline. We include these additional results in the appendix. 

We also compare to two state-of-the-art baselines for training on the iNaturalist dataset: \textbf{LWS}~\cite{kang2020decoupling}, which represents a solution which performs sample re-balancing; and \textbf{LDAM}~\cite{cao2019learning}, which represents a solution which performs loss re-weighting. We chose these two methods because they were state-of-the-art and because they compared against a broad set of other imbalanced learning methods. \textbf{LWS} trains in two stages: first training a standard cross-entropy model with uniform sampling and no re-weighting for 90 epochs; then freezing all layers of the model except for the final classification layer and retraining for 15 epochs using class-balanced sampling. \textbf{LDAM} also uses a two stage training approach: first training a classification model using uniform sampling and weighting with the Label-Distribution Aware Margin (LDAM) loss proposed in their paper for 60 epochs; and then continue training the same model by re-weighting the loss for individual examples based upon their frequency for 30 epochs. 

We train these models using the official code provided by each method~\footnote{LWS: \url{https://github.com/facebookresearch/classifier-balancing}}~\footnote{LDAM: \url{https://github.com/kaidic/LDAM-DRW}}, setting the hyperparameters based on the descriptions in the paper for training on iNaturalist. For \textbf{LWS}, we train the base representation model for the first stage using the same hyperparameters for the \textbf{FT} method when $N\!<\!5089$ (when $N\!=\!5089$, we use the hyperparameters suggested in the paper) and we report the second stage method with the highest performance: for $N\!=\!100$ and $N\!=\!1100$, the \emph{Tau Norm} method; for $N\!=\!5089$, the \emph{LWS} method. 

\begin{table}[t!]
\begin{center}
\small
\begin{tabular}{ lcccccccccc }
\toprule
\multirow{4}{*}{} & \multicolumn{10}{c}{Setting} \\
\cmidrule(lr){2-11}	
& \multicolumn{2}{c}{$N=1$} & \multicolumn{2}{c}{$N=10$} &
\multicolumn{2}{c}{$N=100$} &
\multicolumn{2}{c}{$N=1100$} & \multicolumn{2}{c}{$N=5089$}\\
& \multicolumn{2}{c}{(99.98\%)} & \multicolumn{2}{c}{(99.83\%)} & \multicolumn{2}{c}{(98.30\%)} & \multicolumn{2}{c}{(77.95\%)} & \multicolumn{2}{c}{(0\%)}\\
\cmidrule(lr){2-3} \cmidrule(lr){4-5} \cmidrule(lr){6-7} \cmidrule(lr){8-9} \cmidrule(lr){10-11}
& AP & F1 & AP & F1 & AP & F1 & AP & F1 & AP & F1 \\
\midrule
FT & 10.6 & 10.8 & 9.3 & 8.9 & 38.3 & 38.4 & 50.9 & \textbf{44.8} & 57.7 & 49.9 \\ 
LDAM & - & - & - & - & 24.7 & 21.1 & 41.6 & 37.1 & 57.4 & 55.1  \\
LWS & - & - & - & - & 35.5 & 36.6 & 42.5 & 37.3 & \textbf{60.0} & \textbf{56.9} \\
\midrule
BG Splitting & \textbf{52.9} & \textbf{51.7} & \textbf{44.6} & \textbf{39.4} & \textbf{47.2} & \textbf{40.7} & \textbf{51.4} & 44.7 & 59.9 & 52.5 \\ 
\bottomrule
\end{tabular}
\end{center}
\caption{Average AP and F1 scores of our method and baselines in scenarios with increasing background dominance.
Our approach performs better than fine tuning (\textbf{FT}) in nearly all settings. Our approach is 8.3 mAP points more accurate than state-of-the-art baselines in the $N\!=\!100$ setting, and 42.3 points higher than \textbf{FT} in the extremely imbalanced $N\!=\!1$ setting.  Prior state-of-the-art baselines for imbalanced training (LDAM~\cite{cao2019learning}, LWS~\cite{kang2020decoupling}) perform worse than \textbf{FT} in background dominated settings.}
\label{tab:across_configs}
\end{table}

\subsection{Comparison with baselines}


\textbf{Comparison to fine-tuning with cross entropy.}
Table~\ref{tab:across_configs} compares AP and F1 scores of our approach with those of \textbf{FT}.
When the background category is less dominant ($N=1100$ and $N=5089$),
our approach performs slightly better than \textbf{FT}.
However, our approach increasingly outperforms the \textbf{FT} baseline as background imbalance is increased ($N\leq100$). 
In the extreme ($N=1$) case, our method still produces a useful model,
beating \textbf{FT} by 42.3 AP points and 40.9 F1 points. 
Although the overall performance of our models drops with decreasing $N$,
it does so gracefully without modifying key hyperparameters
(e.g., learning rate, batch size, $b_0$, $\lambda_G$).
Hyperparameter robustness across different levels of background
imbalance is an attractive property of our method that makes it easy to use in practice.

We note that the category distribution of ImageNet is not a superset of the category distribution of iNaturalist (see Figure~\ref{fig:auxloss}) and we did not do any manual labeling or correction of the labels produced by the ImageNet model. Despite this, the pseudo-labels result in a drastically improved feature representation (compared to training with just the main rare category classification loss) that yields higher performance on the foreground categories. 

\textbf{Comparison to LDAM and LWS.}
Table~\ref{tab:across_configs} also compares our method's performance against \textbf{LDAM} and \textbf{LWS}.
The configuration $N\!=\!5089$ is the standard iNaturalist training setup these methods were designed for (no background category), and in this regime, \textbf{LWS} achieves the highest overall AP and F1 score,
and \textbf{LDAM} achieves the second highest F1 score.
However, in the $N\!=\!1100$ and $N\!=\!100$ setting, both methods degrade rapidly, performing worse than the \textbf{FT} baseline, even when using the best hyperparameter configurations found for each $N$.
Class-balanced sampling (\textbf{LWS}) or loss re-weighting based on class frequency (\textbf{LWS}) cause accuracy on the dominant background category to decrease, which directly results in a significant increase in false positives for the foreground categories. This results in a minor increase in the recall for the foreground categories, but a drastic reduction in precision and overall worse performance for \textbf{LDAM} and \textbf{LWS} in the sparse positive setting. Since the performance of \textbf{LDAM} and \textbf{LWS} continues to degrade as $N$ becomes smaller, we do not evaluate these methods for $N=10$ and $N=1$ to save experimental costs.

\begin{table}[t!]
\begin{center}
\small
\begin{tabular}{lll}
\toprule
Classifier Method & Feature Representation & mAP \\
\midrule
Binary SVM & ImageNet & 25.0 \\
& Aux Loss & 28.4 \\
\midrule 
FT & ImageNet & 25.5 \\
& Aux Loss & 28.3 \\
& ImageNet (not frozen) & 38.3 \\
\midrule
BG Splitting & ImageNet (not frozen) & 47.2 \\
\bottomrule

\end{tabular}
\end{center}
\caption{Analysis of the features learned using the auxiliary loss (Aux Loss) versus the base ImageNet features (ImageNet). The Aux Loss features improve over the baseline ImageNet features when using either an SVM or standard SGD to train the final classification layer. The performance increase from fine-tuning the features (the "not frozen" methods is significant, indicating that adjusting the features for the target foreground categories is essential to achieve high performance.} 
\label{tab:aux_task}
\end{table}

\subsection{Auxiliary loss feature analysis}
\label{sec:aux_task_analysis}
How necessary is it to jointly learn the main loss and auxiliary loss? Can the auxiliary loss alone produce an effective feature representation without using any target foreground category labels? If this was the case, one could use such a feature representation to more easily adapt to new foreground categories.
To answer these questions, we conducted an experiment in which we trained the feature representation for the network using only the auxiliary loss, froze the network weights up to the final classification layer, and then trained the final layer using both the main loss with SGD and standard linear SVM training. This configuration decouples the feature representation updates due purely to the auxiliary loss from the contributions of the joint loss. Table~\ref{tab:aux_task} shows the results for the $N=100$ setting, including two end-to-end trained (i.e., without frozen weights) methods for comparison. Even though removing foreground labels keeps 98.3\% of the labels the same, the resulting features are significantly worse (FT-Aux Loss, row 4 in Table~\ref{tab:aux_task}), dropping mAP by 18.9\%. These results show that jointly fine-tuning the feature representation with the auxiliary loss \emph{and} the main loss--which makes use of labeled examples for the target rare categories--is critical, despite the small number of positive examples provided by the dataset for the rare categories.



\begin{table}[t!]
\begin{center}
\small
\begin{tabular}{ lrrrrrrrrrrrr}
\toprule
& \multicolumn{2}{c}{None (\textbf{FT})} & \hspace{.1in} & \multicolumn{2}{c}{+ Aux Loss} & \hspace{.1in} & \multicolumn{2}{c}{+ BG thresholding} &\hspace{.1in} & \multicolumn{3}{l}{
+ Both (\textbf{BG Splitting})} \\
 \cmidrule(lr){1-13}
  & AP & F1 &   & AP & F1 &   & AP & F1 &   & AP & F1\\
  \midrule
& 36.0 & 35.6 &  & 46.0 & 40.4 &  & 41.1 & 37.7 &  & 47.2 & 40.7 \\
\bottomrule
\end{tabular}
\end{center}
\caption{Factor analysis analyzing different components of our approach ($N\!=\!100$, batch size 1024).
Individually, auxiliary loss and background clamping improve the baseline's performance,
but the majority of the benefit comes from the use of the auxiliary loss.} 
\label{tab:ablation_method}
\end{table}

\subsection{Auxiliary loss synergy with other background imbalance methods} Since the auxiliary loss can be easily combined with other background imbalance methods which modify the main classification loss, we are interested in whether the auxiliary loss combined with other methods leads to net improvements in performance. If so, it would mean that the auxiliary loss could be dropped into any existing background imbalance method. We evaluate the benefit by performing a factor analysis to understand how the auxiliary loss combines with the BG thresholding method to produce our total mAP improvement in the $N=100$ setting (Table~\ref{tab:ablation_method}). The "+ Aux loss" only employs the auxiliary loss, while the "+ BG thresholding" only uses the BG thresholding method. We find that both of these methods provide boosts of 10.0 and 5.1 mAP points independently. However, together they produce a net improvement of 11.2 mAP points (+ Both) over the \textbf{FT} baseline. This result shows that the auxiliary loss is a complementary method to other background imbalance methods.

\section{Conclusion}

We discover that state-of-the-art classification methods for handling imbalanced data do not perform well in the presence of a dominant and diverse background.
In response we contribute a new approach that reduces the imbalance by jointly training the main classification task with an auxiliary classification task for categories that are less rare.
This approach can improve on baselines by over 42 AP points in situations of extreme imbalance,
making it feasible to train useful binary classification models for rare categories from only a small number of positives.
We also contribute the iNaturalist-BG datasets, which we hope will encourage further research in this important model training regime.   
\section*{Broader Impact}

Finding rare categories in a sea of background is a common problem encountered in many real-world efforts to leverage visual recognition.
For example, in domains such as medical imaging, autonomous vehicle development, and personal assistant technology, critical categories of interest are rare (maladies, dangerous highway situations, and interesting daily interactions, etc.).
We believe it is important to point the community toward this task,
which is not well served by existing benchmarks, datasets, or methods.  We believe progress will significantly expand the utility of machine vision to new domains and fields.

We acknowledge that improving technologies that sift through large datasets for rare categories also has the potential for misuse.  For example, if a category could be used to identify an individual (e.g., identifying a particular bicycle, car, or t-shirt), our work could be used to advanced applications that infringe on individual privacy.

{\small
\bibliographystyle{plainnat}
\bibliography{main}
}

\begin{appendices}
\section{Auxiliary loss sensitivity to choice of pseudo-labels}

\begin{table}[t!]
\begin{center}
\small
\begin{tabular}{lcc}
\toprule
\multirow{2}{*}{Auxiliary Labels} & \multicolumn{2}{c}{BG Splitting} \\
\cmidrule(lr){2-3}
& AP & F1 \\
\midrule
None & 41.1 & 37.7 \\
Random & 37.2 & 35.0 \\ 
Places &  39.3 & 34.1\\
Cluster-1k & 45.9 & 39.0\\ 
Cluster-5k & 45.9 & \textbf{41.0}\\ 
ImageNet & \textbf{47.2} & 40.7 \\
\bottomrule
\end{tabular}
\end{center}
\caption{Performance of our approach with different sources and methods for supervising the auxiliary loss. Using labels from Places model trained for scene classification on a significantly different distribution hurts performance, whereas using a model trained for ImageNet classification which has overlap with iNaturalist significantly improves performance.}
\label{tab:auxiliary_labels}
\end{table}
To understand how sensitive the auxiliary loss is to the choice of pseudo-labels used as supervision for training the loss, we explored how the choice of different methods for generating these pseudo-labels influenced the final model performance. Our expectation is that using pseudo-labels generated by a process which is more "similar" to the iNaturalist label structure should produce better performance improvements. More specifically, we explore pseudo-labels generated (roughly ordered from least to most similar to iNaturalist labels): at random; using a model pre-trained on a very different task (Places 365~\cite{zhou2017places}); from using clustering algorithms on features extracted from iNaturalist images using a pre-trained ImageNet model; and from using the ImageNet pseudo-labels as in all experiments in the rest of the paper. Table~\ref{tab:auxiliary_labels} summarizes these results for $N\!=\!100$ and batch size 1024. \textbf{None} corresponds to a model trained with no auxiliary loss but with background thresholding. For \textbf{Random},  labels are assigned randomly to each image among 1000 different categories. As expected, the \textbf{Random} label assignment--which ensures category skew but not visual or semantic coherence within the categories--degrades model performance compared to using no auxiliary loss. \textbf{Places} corresponds to generating auxiliary labels using a model pre-trained on the Places365~\cite{zhou2017places} dataset. Note that the Places365 dataset is a scene classification dataset. Using Places365 labels is better than \textbf{Random}, but still degrades performance compared to not using the auxiliary loss because Places365 is a drastically different task. \textbf{ImageNet} corresponds to using auxiliary labels that are generated using a model pre-trained on ImageNet. Since ImageNet has a significant number of animal instances, there is substantial improvement compared training with no auxiliary labels (\textbf{None}). Instead of directly using the labels from an ImageNet pre-trained model, we can use the embeddings produced by the ImageNet model to assign labels by clustering the embeddings into a fixed number of clusters. Both \textbf{Cluster-1k} and \textbf{Cluster-5k} correspond to generating the labels by clustering the embeddings of all the iNaturalist training data computed with an ImageNet pre-trained model using 1k and 5k clusters respectively. We used a fast and approximate version of K-means clustering~\footnote{\url{https://scikit-learn.org/stable/modules/generated/sklearn.cluster.MiniBatchKMeans.html}} to compute the clusters. The labels generated with clustering also give a significant boost compared to either \textbf{None} or \textbf{Random}. The \textbf{ImageNet} pseudo-labels provide the highest AP gain, whereas the clustering generated labels provide the highest F1 gain.

\section{Generalization of BG Splitting learned features}

\begin{table}[t!]
\begin{center}
\small
\begin{tabular}{lcc}
\toprule
Method & Weight Initialization & mAP on 100 new categories (S2) \\
\midrule
Fine-tune last layer & FT on 100 categories (S1) & 15.4 \\
Fine-tune last layer & BG Splitting on 100 categories (S1) & 32.4 \\
\midrule
\multicolumn{2}{l}{BG Splitting (full model training on S2 categories)} & 44.4 \\
\bottomrule
\end{tabular}
\end{center}
\caption{Feature generalization study. Using BG Splitting to train a model for one set of categories (S1) results in features than generalize well for a different set of categories (S2) compared to FT.}
\label{tab:transfer_study}
\end{table}

Are the features learned with Background Splitting more useful for training a classifier for a new set of categories than the features learned from a standard fine-tuning baseline? That is, are the features learned by BG Splitting just better for classifying the target set of categories, or does BG Splitting produce features which can be effectively adapted to classify new categories? Table~\ref{tab:transfer_study} shows the accuracy of a linear classifier trained to classify a set of 100 new classes S2 using features produced by two methods trained on a different set of 100 classes S1: a standard fine-tuning baseline (FT), and our method (BG Splitting). The bottom row is the result of training the BG Splitting method to directly classify the S2 classes and thus represents an upper bound. We find that the features produced by the BG Splitting method are very useful for classifying new sets of categories, as seen by the 17 point mAP increase from using the BG Splitting features versus the FT features. Note that there is still significant advantage to training the entire model for the target set of new categories S2, as seen by further 12 point boost from 32.4 to 44.4 mAP, echoing our observation in Section~\ref{sec:aux_task_analysis} that using the small set of positives for the foreground categories is critical.

\section{Batch size study}
\label{sec:batch_size_study}
\begin{table}[t!]
\begin{center}
\small
\begin{tabular}{rcccccccc}
\toprule
\multirow{2}{*}{Batch size}& \multicolumn{4}{c}{FT} & \multicolumn{4}{c}{BG Splitting} \\
\cmidrule(lr){2-5} \cmidrule(lr){6-9} 
& AP & F1 & Recall & Precision & AP & F1 & Recall & Precision \\
\midrule
128 & 28.4 & 19.0 & \textbf{53.8} & 12.9 & 48.2 & 44.8 & \textbf{53.0} & 44.0\\ 
256 & 37.7 & 30.8 & 50.8 & 24.7 & \textbf{48.8} & \textbf{46.6} & 42.8 & 61.3\\
512 & \textbf{38.3} & \textbf{38.4} & 42.5 & 41.1 & 47.9 & 44.3 & 37.4 & 68.1 \\
1024 & 36.0 & 35.6 & 35.2 & \textbf{44.1} & 47.2 & 40.7 & 33.4 & \textbf{67.1}\\ 
\bottomrule
\end{tabular}
\end{center}
\caption{Performance of standard fine-tuning and our approach with different batch sizes. Larger batch sizes than the ones traditionally used in balanced training (64 or 128) significantly improves performance of standard fine-tuning. Our approach is more robust to batch size but also benefits from a larger batch size.}
\label{tab:batch_size}
\end{table}

Classification networks are typically trained with a batch size of 64 or 128. However, we empirically observed that, in settings with a large background category, larger batch sizes (512 for traditional fine-tuning and 256 for our approach) result in improved performance over small batch sizes. To understand this effect, we evaluated the performance of both traditional fine-tuning and our approach for different batch sizes (Table~\ref{tab:batch_size}). For traditional fine-tuning, a batch size of 512 results in the best performance. Our approach is more robust to smaller batch sizes and marginally benefits from increasing the batch size from 128 to 256. A clear empirical trend we observed is that increasing the batch size results in higher model precision at the cost of lower recall. Our hypothesis for this trend is that as the batches become larger, the model sees more positives and hard negatives in the same batch, requiring the model to be more discriminative, thus increasing precision at the cost of reducing recall. Further exploration of this behavior could lead to better sampling and batch size selection techniques in scenarios with extreme imbalance and a majority background category. We note that all other experiments in this paper are performed with a batch size of 1024 in order to produce results in a reasonable time--although training with smaller batch sizes results in a small increase in performance, it requires training much longer and with lower learning rates to reach convergence.

\section{Background category downsampling}
\begin{table}[t!]
\begin{center}
\small
\begin{tabular}{ lcccccccc }
\toprule
\multirow{4}{*}{} & \multicolumn{8}{c}{Setting ($N=100$)} \\
\cmidrule(lr){2-9}	
& \multicolumn{2}{c}{1\%} & \multicolumn{2}{c}{5\%} &
\multicolumn{2}{c}{25\%} &
\multicolumn{2}{c}{100\%} \\
& \multicolumn{2}{c}{(36.64\%)} & \multicolumn{2}{c}{(74.30\%)} & \multicolumn{2}{c}{(93.52\%)} & \multicolumn{2}{c}{(98.30\%)}\\
\cmidrule(lr){2-3} \cmidrule(lr){4-5} \cmidrule(lr){6-7} \cmidrule(lr){8-9} 
& AP & F1 & AP & F1 & AP & F1 & AP & F1 \\
\midrule
FT & 23.5 & 18.7 & 27.9 & 22.4 & 36.3 & 32.5 & 38.3 & 38.4 \\ 
LDAM & 23.0 & 5.7 & 14.8 & 3.3 & 13.2 & 2.7 & 24.7 & 21.1 \\
LWS & 15.0	& 3.9 & 12.7	& 3.8 & 38.8 & 39.1 & 35.5 & 36.6 \\
\midrule
BG Splitting & - & - & - & - & - & - & \textbf{47.2} & \textbf{40.7} \\ 
\bottomrule
\end{tabular}
\end{center}
\caption{AP and F1 scores for baseline methods using a downsampled background category. In all cases, using the downsampled background category dataset does not result in performance which exceeds our method. In nearly all cases, the downsampled dataset results in reduced performance because it drastically decreases precision, resulting in a large number of false positives.}
\label{tab:background}
\end{table}

As the training issues caused by the extremely large background category are due to imbalance between the number of foreground and background examples, one might ask if simply 'downsampling' the number of background category instances would solve the problem. We implemented this approach by selecting increasingly smaller fractions of the background category images in the training set, and creating new training sets using all the foreground images plus this smaller background set. Table~\ref{tab:background} shows the AP and F1 score for the baselines when trained with the downsampled background datasets for 100\%, 25\%, 5\%, and 1\% of all background instances. In nearly all cases, using the downsampled versions of the dataset results in lower overall performance across both AP and F1. In particular, we find that using these downsampled datasets increases the recall of the baseline methods, but at drastic cost to precision, resulting in a lower overall F1 and AP score. 



\end{appendices}

\end{document}